\documentclass{article}

\usepackage{makecell}
\usepackage{amsmath,amsfonts,stmaryrd,amssymb} 

\usepackage{enumerate} 

\usepackage[ruled]{algorithm2e} 

\usepackage[framemethod=tikz]{mdframed} 

\usepackage{listings} 
\usepackage{geometry} 

\usepackage[utf8]{inputenc} 
\usepackage[T1]{fontenc} 

\usepackage{XCharter} 

\mdfdefinestyle{commandline}{
	leftmargin=10pt,
	rightmargin=10pt,
	innerleftmargin=15pt,
	middlelinecolor=black!50!white,
	middlelinewidth=2pt,
	frametitlerule=false,
	backgroundcolor=black!5!white,
	frametitle={Command Line},
	frametitlefont={\normalfont\sffamily\color{white}\hspace{-1em}},
	frametitlebackgroundcolor=black!50!white,
	nobreak,
}

\mdfdefinestyle{file}{
	innertopmargin=1.6\baselineskip,
	innerbottommargin=0.8\baselineskip,
	topline=false, bottomline=false,
	leftline=false, rightline=false,
	leftmargin=2cm,
	rightmargin=2cm,
	singleextra={%
		\draw[fill=black!10!white](P)++(0,-1.2em)rectangle(P-|O);
		\node[anchor=north west]
		at(P-|O){\ttfamily\mdfilename};
		\def\l{3em}
		\draw(O-|P)++(-\l,0)--++(\l,\l)--(P)--(P-|O)--(O)--cycle;
		\draw(O-|P)++(-\l,0)--++(0,\l)--++(\l,0);
	},
	nobreak,
}

\mdfdefinestyle{question}{
	innertopmargin=1.2\baselineskip,
	innerbottommargin=0.8\baselineskip,
	roundcorner=5pt,
	nobreak,
	singleextra={%
		\draw(P-|O)node[xshift=1em,anchor=west,fill=white,draw,rounded corners=5pt]{%
		Question \theQuestion\questionTitle};
	},
}

\newcounter{Question} 

\mdfdefinestyle{warning}{
	topline=false, bottomline=false,
	leftline=false, rightline=false,
	nobreak,
	singleextra={%
		\draw(P-|O)++(-0.5em,0)node(tmp1){};
		\draw(P-|O)++(0.5em,0)node(tmp2){};
		\fill[black,rotate around={45:(P-|O)}](tmp1)rectangle(tmp2);
		\node at(P-|O){\color{white}\scriptsize\bf !};
		\draw[very thick](P-|O)++(0,-1em)--(O);
	}
}

\mdfdefinestyle{info}{%
	topline=false, bottomline=false,
	leftline=false, rightline=false,
	nobreak,
	singleextra={%
		\fill[black](P-|O)circle[radius=0.4em];
		\node at(P-|O){\color{white}\scriptsize\bf i};
		\draw[very thick](P-|O)++(0,-0.8em)--(O);
	}
}

\title{Investigating Plausibility of Biologically Inspired Bayesian Learning in ANNs} 
\author{Ram J. Zaveri  \\ \texttt{rz0012@mix.wvu.edu}}

\date{West Virginia University} 

\begin{document}

\maketitle 

\section{Abstract}
\hspace{\parindent}\textit{Catastrophic forgetting has been the leading issue in the domain of lifelong learning in artificial systems. Current artificial systems are reasonably good at learning domains they have seen before; however, as soon as they encounter something new, they either go through a significant performance deterioration or if you try to teach them the new distribution of data, they forget what they have learned before. Additionally, they are also prone to being overly confident when performing inference on seen as well as unseen data, causing significant reliability issues when lives are at stake. Therefore, it is extremely important to dig into this problem and formulate an approach that will be continually adaptable as well as reliable. If we move away from the engineering domain of such systems and look into biological systems, we can realize that these very systems are very efficient at computing the reliance as well as the uncertainty of accurate predictions that further help them refine the inference in a life-long setting. These systems are not perfect; however, they do give us a solid understanding of the reasoning under uncertainty which takes us to the domain of Bayesian reasoning. We incorporate this Bayesian inference with thresholding mechanism as to mimic more biologically inspired models, but only at spatial level. Further, we reproduce a recent study on Bayesian Inference with Spiking Neural Networks for Continual Learning to compare against it as a suitable biologically inspired Bayesian framework. Overall, we investigate the plausibility of biologically inspired Bayesian Learning in artificial systems on a vision dataset, MNIST, and show relative performance improvement under the conditions when the model is forced to predict VS when the model is not.}

\section{Introduction}

\hspace{\parindent}In recent years, we have encountered many problems with Aartificial Neural Networks (ANNs), for example, these very ANNs, even though, are pretty good approximators; they fall short when they meet information that does not belong to their prior distributions, and therefore; go through significant performance deterioration. Biological systems are well equipped to handle that. The population modeling of the neuronal dynamics suggests that each neuron possesses a temporal and spatial threshold to process the information flow to the next neuron. This essentially helps determine “if” the neuron should be activated by the prior input at this very time instance. If there is an accumulation of ions in the pre-synaptic cleft, the next neuron is more likely to get activated to process the signal ~\cite{barrett2020seven}. I believe this is the time we go back to the biological drive to develop more efficient artificial systems. The reason is, that threshold-based mechanisms, at the very high level at least, help the neuron decide the functional co-dependence of the neuron’s activation and the signals it is processing.  \\

In the case of ANNs, parameters such as weight and bias help it refine the distribution clusters in the hyperspace. The way the calculations are performed is through the accumulation of weights from previous neurons passed through an activation function (ReLU, Sigmoid, etc. ) to add the nonlinearity to the prior layer, and then process it to the next layer. Of course, activation layers are extremely important since they add non-linearity to the neural network and help with exploding weights from the previous layer. However, it is assumed that once the model has been trained, the learned parameters, weights, and biases, are optimal to the unseen data. This is not the case in general, the ANNs overfit to the seen data, and either do not perform well or if trained again on new data, they will experience catastrophic forgetting. In essence, we need to add a means of certainty in the learning process. The other way to say this is to add a level of uncertainty to individual neurons. Scientists are trying to tackle this problem by defining means and variance of the weights; it is referred to as Bayesian Neural Networks (BNNs). The goal here is to learn the certainty of the accumulated weights based on the firing pattern ~\cite{Jaynes_2003}. \\

As described in ~\cite{Skatchkovsky_2022}, the free energy hypothesis suggests measuring surprise in terms of variational free energy. If we put it in the context of the biological brain, it constantly updates an internal model of the world and tries to minimize the information-theoretic surprise, it is formerly known as the Bayesian brain hypothesis.  In essence, synaptic plasticity can be formulated as a Bayesian learning mechanism where it keeps track of, not the weights, but the distribution of synaptic weights over time ~\cite{Aitchison_2021, Friston_2012}. Therefore, investigating the probability of an event occuring, rather than saying that this particular event has occurred, would give us a more generalized, and symmetrical point of view on the problem of continual adaptibility. In essense, the goal here is to predict a continuous variable to determine certainty of an event would give a more biologically plausible answer. \\

Widely understood mechanisms that describe spike-time plasticity, long-term potentiation (LTP), and long-term depression (LTD) follow similar concepts, where LTP occurs when the neurons spike in sequence to strengthen the synapse between two consecutive neurons, essentially experiencing an increase in weight; on the other hand, LTD does the opposite to weaken the synapses, essentially decreasing the weight. LTP contributes to long-term memory by consolidating synaptic connections, forming stability throughout the network. It occurs via either episodic replay of sequential pre-synaptic signals (repeating) or if the pre-synaptic signals were extremely strong (high-frequency) to induce an LTP. This essentially means that LTP can be caused by either temporal summation or spatial summation or both~\cite{McEachern_1999}. In biological networks, narrowly tuned Ca\(^{+2}\) release is known to be associated with different types of LTP, usually causing short-term potentiation (STP) for minutes to hours to cause late LTP ~\cite{Raymond_2002, Cao_2012}. However, if the network has experienced a similar situation before, it will cause it to go through a synaptic change called metaplasticity ~\cite{Crestani_2018}. Moreover, during an inactive state (sleep), these very connections become more refined by episodic replay ~\cite{soures2021tacos}. We will leverage this information and design a paradigm for the artificial systems using learnable activation/thresholds as a means of spatial summation. Additionally, we will use terms activation/thresholds interchangeably throughout this paper. \\    
\section{Related Works}

\hspace{\parindent} Bayesian Neural Networks are increasingly gaining popularity in deep learning community, where more recently, \cite{jospin2022hands} combines the most recent research to prepare a tutorial specifically for deep learning practitioners. The advantage of BNNs is that they are not prone to overfitting as can be observed in traditional learning schemes. They are being used in many fields, i.e., computer vision \cite{kendall2017uncertainties}, network traffic monitoring \cite{auld2007bayesian}, medicine \cite{beker2020minimal}, active learning \cite{gal2017deep, tran2019bayesian}, online learning \cite{opper1999bayesian}, and so on. Additionally, as observed in \cite{ritter2018online}, since the calculated posteriors can be reused as priors for the data the model has not seen before, it can inherently avoid the major problem of catastrophic forgetting. Other key advantages are they naturally learn to quantify uncertainty, they can even distinguish between uncertainty associated with the data it has seen and the data it has never seen, making it robust against anomaly \cite{depeweg2018decomposition}. Additionally, they are also different from, so called, black box algorithms with traditional deep learning models since, the prior that are used in BNNs are explicit \cite{jospin2022hands}. \cite{khan2017conjugate, khan2021bayesian} further incorporates Bayesian Learning rule with natural gradient descent in a fashion it works, if not exactly, similarly, without destroying the core fundamentals of machine learning principles, with machine learning algorithms. \\

Continual learning refers to a setting where an algorithm is learning in a continuous fashion. Thus said, the algorithm does need to preserve the past information it learned; however, most approaches struggle in this particular setting and go through catastrophic forgetting \cite{ritter2018online}. As we mentioned earlier, BNNs are stochastic models and learn continuous distributions (one can also discretize this \cite{Skatchkovsky_2022}), probabilistic approximations are more stable compared to their ANNs counterparts. Additionally, there are tremendous research works mentioning various mathematical and empirical tricks, i.e., meta learning \cite{hospedales2021meta}, regularization \cite{ahn2019uncertainty}, etc., to avoid catastrophic forgetting; however, none have made any significant breakthroughs. On the other hand, more on the biologically plausible models, which are less discrete, \cite{laborieux2021synaptic,ritter2018online}, have made progress in terms of conceptual significance of Spiking Neural Netowkrs (SNNs), specifically leaky integrate-and-fire (LIF) model, on standard continual learning benchmarks. Here, SNNs are wonderful tools as they can be mimicked at both, software and hardware-level, which is to show the significance of memory and energy-efficiency, which we call neuromorphic tools. Here, our major focus \cite{Skatchkovsky_2022}, have proposed to use Bayesian Learning with SNNs in a continual fashion. They use variational inference with surrogate gradients to perform Bayesian learning in SNNs, and are achieving noticeable performance gain. In this paper, we are investigating a rather simplistic model with learnable activations to showcase the behavior of Bayesian framework on unforced certainty prediction and forced prediction. We further incorporate \cite{Skatchkovsky_2022} framework to achieve extreme biological plausibility. We note that, we are thoroughly investigating Bayesian Learning in this paper to provide better conceptual understanding and not focusing on the basics of LIF spiking models since the class has covered major portions of that, the other mechanisms remain the same for the most part. 
\section{Methods}

\subsection{Overview}
\hspace{\parindent} We investigate the plausibility of biologically inspired Bayesian Learning in ANNs by first investigating the core componets of Bayesian Neural Netowkrs (BNNs). We further describe thresholding as ANNs counterpart to biological activation of a neuron. We further discuss its applicability in a continual learning situation integrated with a more biologically plausible model, Spiking Neural Networks (SNNs) from \cite{Skatchkovsky_2022}.

\subsection{Bayesian Neural Networks}\label{sec:BNN}

\hspace{\parindent} As opposed to the frequentist approach, Bayesian Inference considers the probability as a measure of belief in the occurrences of events rather than the limit on the frequency, meaning the hypothesis testing is done on the probability distribution of events occurring, which again brings us to the theory of variational free energy that describes the probabilistic model to predict observations from the hypothesized causes \cite{jospin2022hands, bruineberg2018anticipating, etz2018become}. Therefore, in Bayesian Inference, the prior beliefs influence the posterior beliefs. See equation \ref{eq:bayes}.
\begin{equation}\label{eq:bayes}
    P(H|D) = \frac{P(D|H)P(H)}{P(D)} = \frac{P(D,H)}{\int_{H}^{}  \,P(D,H')dH'}  
\end{equation}
Where $P(D|H)P(H)$ is the likelihood, $P(H)$ is the prior, $P(D)=\int_{H}^{}  \,P(D,H')dH'$ is the evidence from the data $D$, $P(H|D)$ is the posterior, and $P(D,H)$ is the joint probability. \\

As mentioned in \cite{jospin2022hands}, BNNs are referred to as stochastic artificial neural networks using Bayesian Inference. Here, stochastic models utilize probability distributions rather than a point estimate of the value unlike the traditional neural networks. This gives a better understanding of precision, or uncertainty, of the weight vector that is associated with every neuron. This is typically done by incorporating a stochastic activation or weight as observed in BNNs. Therefore, they have capacity to simulate various possible solutions, inherently forming an ensemble of networks. For example, assume $\bf{\theta=(W,b)}$, where $\theta$ is the model parameters, $\bf{W}$ and $\bf{b}$ are weights and biases, respectively. The traditional networks will learn through a point estimate equation, $s(\bf{Wl+b})$, where $s$ is the activation function. In stochastic models, either $s$ or $\bf{\theta=(W,b)}$ are associated with probability distributions. And since the computation is inherently dependent on the probability distributions of either $s$ or $\theta$, stochastic models are forming ensembles of distributions, thus, can be used as a special case of ensemble learning \cite{zhou2012ensemble}. \\

To train the BNNs, we will first have to establish the Bayesian Posterior:
\begin{equation}\label{eq:bayes_post}
    p(\theta|D) = \frac{p(D_y|D_x, \theta)p(\theta)}{\int_{\theta}p(D_y|D_x, \theta')p(\theta')d\theta'}  \varpropto  p(D_y|D_x, \theta)p(\theta)
\end{equation}
During prediction for a given $p(\theta|D)$, we can compute the prediction, $p(y|x,D)$ (sampled from $y = \Phi_\theta(x) + \epsilon $, where $\Phi$ is an approximation and $\epsilon$ is the noise) as follows:
\begin{equation}\label{eq:bayes_post}
    p(y|x,D) = \int_{\theta}p(y|x, \theta')p(\theta'|D)d\theta'
\end{equation}
Here, computing $\int_{\theta}p(D_y|D_x, \theta')p(\theta')d\theta'$, is extremely difficult and time-consuming, since we are trying to perform integral over all the possible parameters, extremely in-efficient. Therefore, one of the two major approaches is used in this work, called Variational Inference. The other one is Markov Chain Monte Carlo; which falls short with increasing scalability as opposed to variational inference \cite{blei2017variational}. 

\paragraph{Variational Inference.\\}
\hspace{\parindent}  Variational Inference \cite{blei2017variational} is not exact; however, provides very good approximation with increasing scale. Here, instead of sampling from an exact posterior mentioned in Equation \ref{eq:bayes_post}, we compute a distribution $q_\phi(H)$, namely variational distribution, parameterized by parameters $\phi$. Overall, we want  $q_\phi(H)$ to be as close to the exact posterior $P(H|D)$ as possible. We use Kullback-Leiber (KL) divergence \cite{kullback1951information}  function to measure that based on Shannon's information theory \cite{shannon1948mathematical}. The overall KL-Divergence loss is as follows:
\begin{equation}\label{eq:kl_diver}
D_{KL}(q_\phi||P) =\int_{H}q_\phi(H')log \Bigl( \frac{q_\phi(H')}{P(H'|D)} \Bigr) dH'
\end{equation}
However, KL-Divergence loss still takes $P(H|D)$ into account, and will still need to calculate it; rather, if we simplify the formula, we get the following:
\begin{equation}\label{eq:elbo}
    log(P(D)) - D_{KL}(q_\phi||P) =\int_{H}q_\phi(H')log \Bigl( \frac{q_\phi(H')}{P(H')} \Bigr) dH'
\end{equation}
Here, during gradient calculation, $log(P(H'|D))$ becomes $log(P(H',D)) - log(P(D))$, where $log(P(D))$ is a constant and becomes 0. Therefore, we can safely eliminate this term and reduce it to the equation \ref{eq:elbo}, which is also referred to as evidence lower bound ELBO loss and the way to optimize this loss is called stochastic variational inference \cite{hoffman2013stochastic}. 

Now, to perform backpropagation over a neural network, we need to calculate gradients over equation \ref{eq:elbo} according to the following equation:
\begin{equation}\label{eq:derivative}
    \frac{\partial}{\partial\phi}\int_{\phi}q_\phi(\theta')f(\theta' ,\phi)d\theta' = \int_{\epsilon} q(\epsilon) \Bigl( \frac{\partial f(\theta,\phi)}{\partial \theta } \frac{ \partial \theta}{ \partial \phi} + \frac{\partial f(\theta,\phi)}{ \partial \phi}  \Bigr)
\end{equation}
where, $q_\phi(\theta)\partial \theta = q(\epsilon)\partial \epsilon$. \\

Further, to estimate the priors during inference, we first need to establish that equation \ref{eq:elbo} is KL-divergence of $q(\phi)$ substracted from the log-likelihood of the data, $log(P(D)) - D_{KL}(q_\phi||P)$, which means equation \ref{eq:elbo} is a function of both the variational parameters, $\phi$, to estimate the posterior, and (addition of ) a parameterized prior distribution, $p_\xi(H)$, given the parameters, $\xi$. Therefore, the loss function now becomes, 
\begin{equation}\label{eq:overall_loss}
    L = log(q_{\phi}(\theta)) - log(p_{\xi}(D_y|D_x, \theta)p_{\xi}(\theta)) 
\end{equation}
Here, we perform gradient decent with respect to both $\phi$ and $\xi$ using the equation \ref{eq:derivative} (replace $\phi$ with $\xi$).

\subsection{Thresholding} \label{sec:thresh}
\hspace{\parindent}  As our baseline comparison, we follow the guidelines provided in section \ref{sec:BNN}. We further describe the stochastic models that make use of the probabilistic parametric functions that complies with the concept of adding a variational threshold for activation for introducing nonlinearity in a stochastic fashion. We diverge from this idea, and test a simple threshold-based mechanism, which takes learnable threshold combined with ReLU as the activation function. Here, the basic idea is to consider spatial information as a means of collective-ness, in-short, threshold. This works as masking of the neurons based on the training data. Consider $s(\bf{Wl+b})$ terms where $\bf{\theta=(W,b)}$ are learened parameterized distributions and $s$ is learnable activation/threshold. Here, we should note that, this thresholding is not a probability distribution just yet, it is a means to provide weight of the neuron given the experience during training, results are shown in section \ref{sec:res}. So far, we can call this scheme as learnable activations. Traditional neural netoworks have shown some performance gain with this scheme, but not significant \cite{goyal1906learning}. However, with Bayesian Learning, it turns out, when the model was forced to predict, the overall accuracy decreased, but surprisingly, when the model was not forced, rather, skipped some inputs when it was "uncertain", the accuracy jump was relatively high, suggesting with learnable activations, the model might be plastic to change, but is relatively stable towards the things it understands. \\

\subsection{Continual Learning} \label{sec:cont_Lear}
\hspace{\parindent}  To test the BNNs with baseline thresholding in a continual learning situation, we choose MNIST \cite{lecun1998mnist} dataset that consists of 10 digits, and following \cite{Skatchkovsky_2022} method, and split them in 5 groups ([0,1], [2,3], [4,5], [6,7], [8,9]) sequentially. We train the model with each group once and in sequence and calculate the performance in terms of accuracy for this particular task.  The results are shown in section \ref{sec:res}. This test was designed to test our baseline thresholding. However, if we incorporate more biologically inspired models, such as leaky integrate-and-fire (LIF), we need to go into the realm of Spiking Neural Networks (SNNs) \cite{soures2021tacos}. But that is not the only motivation, learnable thresholding with the temporal component associated with it would result in a more biologically plausible Bayesian Learning framework. This is an inherent property of the SNNs. Therefore, it is natural to investigate these spiking models for continual learning, specifically as a Bayesian Learning framework.  \cite{Skatchkovsky_2022} tackles this and achieves remarkable performance. We incorporate their framework and test it against previously described mechanisms, and show the results in section \ref{sec:res}.

\section{Results} \label{sec:res}

\subsection{Implementation Details}
\hspace{\parindent}  All the code was written in Python and the models are implemented in PyTorch \cite{paszke2019pytorch}. All the models in this work are using 784 as input dimension (MNIST \cite{lecun1998mnist} dimensions are 28$\times$28 $\in$ $R^2$  $\to$ 784 $\in$ $R^1$), 400 as hidden dimensions, and 10 as output dimensions given the number of digits in MNIST \cite{lecun1998mnist} are 10. All the nodes are fully connected nodes. Models without spikes are merely traditional MLPs with additional functionality of learnable thresholding or activation. For direct test on all 10 digits, we train the model for 5 epochs, and during continual learning, we only train it for one epoch every task following the implementation of \cite{Skatchkovsky_2022}, i.e., [0,1] category, then [2,3] category, so on and so forth. 

\subsection{Baseline Testing}
\hspace{\parindent}  To investigate biological plausibility of Bayesian Inference, we incorporated learnable threshold/activation to the generic BNNs. Table \ref{table:preliminary} illustrates that under forced conditions generic BNNs achieve significantly better accuracy compared to their counterpart with learnable activations. However, as we incorporate their uncertainty into the prediction and discard values with lower precision than 0.8 for any given class, we see that BNNs with learnable thresholds achieves comparatively good accuracy. We should note that, this is only a preliminary test and is to investigate if the BNNs are compatible with learnable thresholds. This shows that they are indeed compatible, and our speculation is that lower certainty describes their ability to be plastic towards fluctuations and stable only when absolutely certain. However, it still needs more investigation to solidify the study since the approach with learnable activation does skip a large number of samples, which might not be optimal.
\begin{figure}[t!]
    \centering
    \includegraphics[width=0.9\textwidth, height=0.6\textwidth]{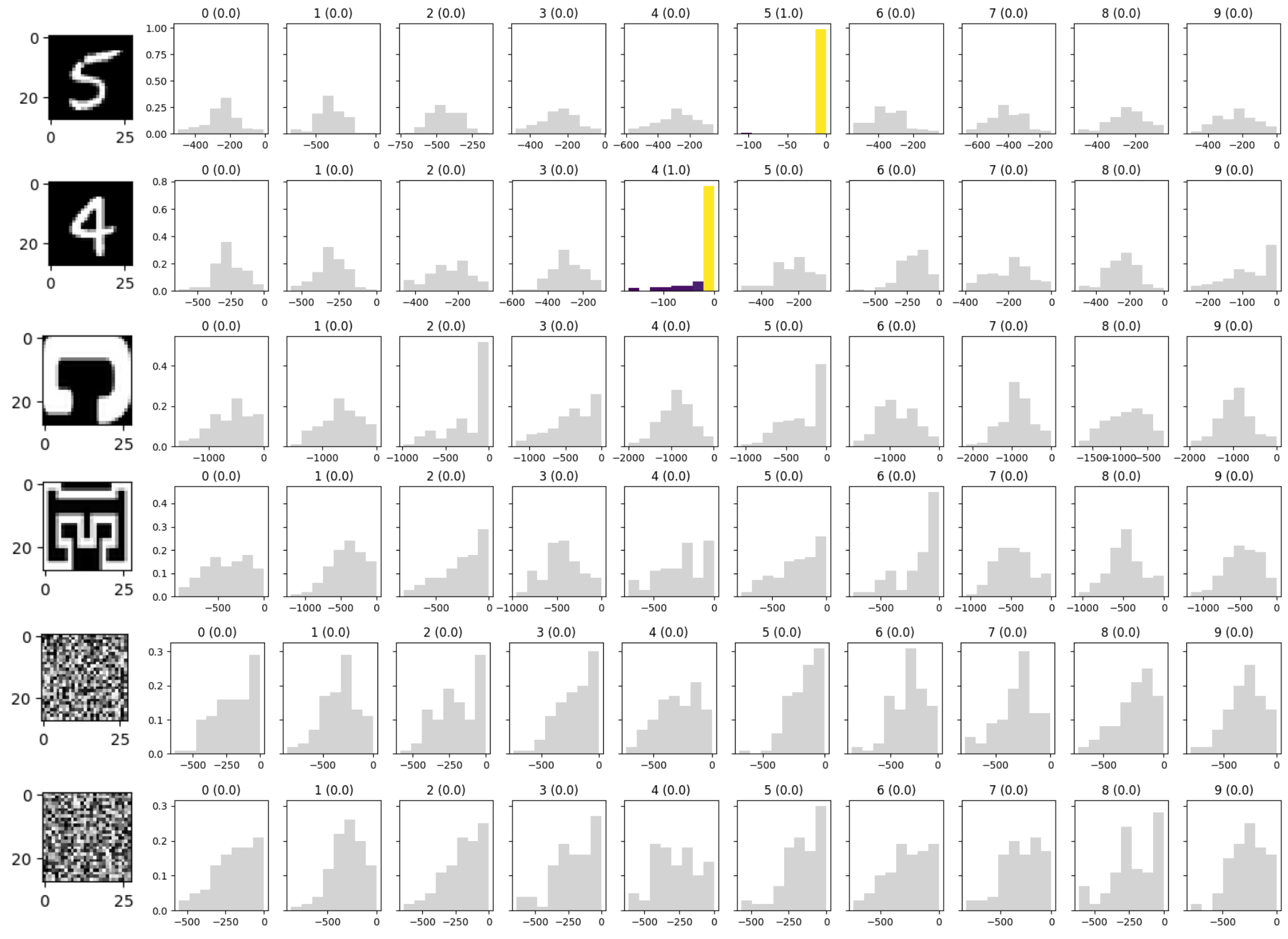}
    \caption{\textbf{Qualitative results on BNNs with Learnable Thresholding.} First two rows are results from the learned classes, the following two rows are from alphabets, (meaningful but) not learned domain, and the last two rows are results with random noice.}
    \label{fig:qual_base}
    \vspace{-5mm}
\end{figure}

\begin{table*}[t!]
    \begin{center}
    \caption{Preliminary Testing when the model is forced to predict VS when the model is not.}
    \label{table:preliminary}
    \scalebox{0.8}{
    \begin{tabular}{c|cc|cc}
    \hline
    & \multicolumn{2}{c|}{Forced Pred.} & \multicolumn{2}{c}{Not Forced Pred.}\\
    & Skipped & Accuracy & Skipped & Accuracy \\
    \hline		
    Genric BNNs & 0/10000 & 0.87 & 1415/10000 & 0.95 \\
    BNNs with Learnable Activations & 0/10000 & 0.79 & 3735/10000 & 0.96 \\
    \hline
    \end{tabular}
    }\end{center}
\end{table*}

\subsection{Testing on Continually changing tasks}

\hspace{\parindent} Table \ref{table:continual} illustrates empirical study on continually changing tasks, specifically on generic BNNs, BNNs with learnable thresholds/activation in conditions where model is forced to predict and when the model is not (skipping predictions are allowed based on the probability), and BNNs with SNNs. Here, this continually changing tasks, as described in section \ref{sec:cont_Lear}, are five sequential tasks split evenly from the MNIST \cite{lecun1998mnist} dataset. As observed from the results, Generic BNNs are performing wonderfully well compared to BNNs with learnable activations in conditions where the model is forced to predict. However, as we can notice, BNNs with learnable activations had a significant improvement when the model was not forced to predict (skipping allowed). We suspect, the reason for that would be the model being more plastic towards the change, whereas the stable connections are maintaining the model performance when the continually learning the tasks. Further, we test BNNs with SNNs (LIF model) based on \cite{Skatchkovsky_2022} study to  showcase the neuromorphic capacity of the BNNs, and the results show comparable improvement over the BNNs without SNNs. Additionally, to get an idea of uncertainty in BNNs qualitatively, please consider Figure \ref{fig:qual_base}.

\begin{table*}[t!]
    \begin{center}
    \caption{Comparative study on Generic BNNs, BNNs with learnable thresholds/activations (with and without skipping allowed), and BNNs with SNNs (LIF model).}
    \label{table:continual}
    \scalebox{0.65}{
    \begin{tabular}{c|c|cc|ccc|cccc|ccccc}
    \hline\noalign{\smallskip}
    & \multicolumn{15}{c}{Tasks} \\
    \hline\noalign{\smallskip}
    Training  & \multicolumn{1}{c|}{Task 1} &   \multicolumn{2}{c|}{Task 2} &   \multicolumn{3}{c|}{Task 3} &   \multicolumn{4}{c|}{Task 4} &   \multicolumn{5}{c}{Task 5}\\
    \hline\noalign{\smallskip}

    & Task 1 & Task 1 & Task 2 & Task 1 & Task 2 & Task 3 & Task 1 & Task 2 & Task 3 & Task 4 & Task 1 & Task 2 & Task 3 & Task 4 & Task 5 \\

    \hline\noalign{\smallskip}	
    Genric BNNs                                                                 & 0.998 & 0.950 & 0.973 & 0.987 & 0.845 & 0.813 & 0.967 & 0.880 & 0.790 & 0.790 & 0.949 & 0.850 & 0.899 & 0.704 & 0.543 \\
    \makecell{ Genric BNNs \\ (skipping allowed)  }                             & 0.999 & 0.978 & 0.968 & 0.997 & 0.901 & 0.902 & 0.965 & 0.883 & 0.840 & 0.930 & 0.976 & 0.896 & 0.895 & 0.842 & 0.526 \\
    \hline\noalign{\smallskip}

    \makecell{BNNs with \\ Learnable Activations}                               & 0.998 & 0.917 & 0.946 & 0.978 & 0.784 & 0.818 & 0.950 & 0.911 & 0.641 & 0.793 & 0.966 & 0.862 & 0.722 & 0.614 & 0.567  \\
    \makecell{BNNs with \\ Learnable Activations \\ (skipping allowed)}         & 0.998 & 0.963 & 0.970 & 0.999 & 0.912 & 0.903 & 0.997 & 0.967 & 0.850 & 0.929 & 0.998 & 0.981 & 0.951 & 0.882 & 0.849  \\
    \hline\noalign{\smallskip}

    \makecell{BNNs with \\ SNNs}                                                & 0.999 & 0.995 & 0.800 & 0.956 & 0.952 & 0.605 & 0.976 & 0.878 & 0.882 & 0.914 & 0.959 & 0.867 & 0.782 & 0.841 & 0.892 \\
    \hline\noalign{\smallskip}
    \end{tabular}
    }\end{center}
\end{table*}

\section{Discussion}
\hspace{\parindent} Since the motivation is to derive a more biologically inspired Bayesian modeling of the network, we investigated, a simple learnable activation/thresholding as can be observed in biological networks in terms of tuning curves. This simple trick helped model learn relevant information in a continual learning situation, where all the data is not present at once, instead, it becomes available sequentially. Further we used study proposed in \cite{Skatchkovsky_2022} and reproduce the results on BNNs with SNNs to show with additional complexity of spiking networks, we can leverage a more biologically inspired Bayesian learning scheme, and shows incredible performance increase. Additionally, our initial motivation was to utilize learnable activation/thresholds in spiking networks; however, upon investigation, it was found that it is their inherent property, spiking rate as measure of significance/activation, and therefore, they are more biologically plausible networks. The results indeed support that. Further, one of the biggest limitations of this work is that the work is still in its preliminary stage, and requires a lot more investigation on simple learnable activation-based BNNs and spiking models. It is quite difficult to draw conclusions based on this study this early; however, that does inspire us to investigate more of their plausibility in the realm of spiking models in the future. Another limitation is the scalability of the network, we are unsure if this particular scheme can be scaled to bigger models and datasets given the complexity of the model is quite limited and also the data. We would like to investigate bigger and more complex models with larger datasets to test the applicability of this work in the real world. Another major limitation is allowing the model to skip certain predictions, the question would be, how many skips are considered okay, we need more investigation on this as well.

\section{Conclusion}
\hspace{\parindent} In this work, we proposed to use learnable thresholding/activation of the neurons as a measure of tuning curves, inspired by biological neurons, to investigate the plausibility of biologically inspired Bayesian learning as mentioned in \cite{Skatchkovsky_2022}, who are using spiking neural networks that work in the LIF fashion. The study still requires more investigation; however, this preliminary study shows that learnable threshold provide a measure of tuning curves and can maintain both stability and plasticity throughout their continual learning tasks when model is not forced to predict. This gives us a new direction of ideas to start thinking in terms of, do the artificial networks really need to make solid decisions, or they can refuse if they are uncertain. Ethically, this could potentially make a huge difference in the field of Artificial Intelligence, and is a wonderful question to ask.  
\section{Acknowledgements}
We appreciate the work shared by \cite{Skatchkovsky_2022} at https://github.com/kclip/bayesian-snn, and a basic tutorial:
https://github.com/paraschopra/bayesian-neural-network-mnist. 
We further thank  Dr. Nicholas Szczecinski for approving this work.

\bibliographystyle{IEEEtran}
\bibliography{main}

\end{document}